\documentclass[10pt,twocolumn,letterpaper]{article}
\usepackage{wacv}
\usepackage{amsmath,amssymb}
\usepackage{booktabs}
\usepackage{graphicx}
\usepackage{multirow}
\usepackage{microtype}
\usepackage{enumitem}
\usepackage{xcolor}
\ifdefined\WACVRevisionMarkup
  \newcommand{\rev}[1]{\textcolor{red}{#1}}
\else
  \newcommand{\rev}[1]{#1}
\fi
\usepackage{pifont}
\usepackage{siunitx}
\usepackage{subcaption}
\usepackage{tabularx}
\usepackage{adjustbox}
\usepackage{float}
\usepackage{placeins}

\newcommand{\cmark}{\ding{51}}
\newcommand{\xmark}{\ding{55}}

\newcommand{\C}{\ensuremath{\mathrm{C}}}
\newcommand{\M}{\ensuremath{\mathrm{M}}}
\newcommand{\CM}{\ensuremath{\mathrm{CM}}}

\setlist[itemize]{leftmargin=*,topsep=2pt,itemsep=1pt,parsep=0pt}
\setlist[enumerate]{leftmargin=*,topsep=2pt,itemsep=1pt,parsep=0pt}
\usepackage{tikz}
\usetikzlibrary{arrows.meta,positioning}
\definecolor{wacvblue}{rgb}{0.21,0.49,0.74}
\usepackage[pagebackref,breaklinks,colorlinks,allcolors=wacvblue]{hyperref}
\def\confName{WACV}
\def\confYear{2027}
\newcommand{\benchmark}{\textsc{LandMarkVLN}}
\title{Do Navigation VLA Models Ground What They Are Told?\\
An Empirical Audit of Landmark-Vision Alignment in VLN}
\author{%
Josh Qixuan Sun$^{1}$ \quad Xiaoying Xing$^{2}$ \quad Yurun Chen$^{3,4}$\\
Huaiyuan Weng$^{1}$ \quad Wentao Zhu$^{4}$ \quad Mark Crowley$^{1}$\\
$^{1}$University of Waterloo \quad $^{2}$Alibaba Group\\
$^{3}$Shanghai Jiao Tong University \quad $^{4}$Eastern Institute of Technology, Ningbo\\
{\tt\small \{josh.q.sun,huaiyuan.weng,mark.crowley\}@uwaterloo.ca,\quad
xiaoyingxy0505@gmail.com}\\
{\tt\small euron.zc@sjtu.edu.cn,\quad wtzhu@eitech.edu.cn}%
}
\hypersetup{pdfauthor={Josh Qixuan Sun, Xiaoying Xing, Yurun Chen,
Huaiyuan Weng, Wentao Zhu, Mark Crowley}}

\begin{document}
\maketitle
\begin{abstract}
Vision-and-language navigation (VLN) requires models to connect route
instructions with visual observations when selecting actions, but standard
navigation metrics do not reveal which visual evidence supports those actions.
We introduce \benchmark{}, a 7,228-sample evaluation dataset constructed from
VLN benchmarks. Each sample contains a navigation instruction, a
\rev{landmark phrase naming one physical object}, current and historical observations, same-instance landmark
masks, and local reference actions. Using \benchmark{}, we audit eight released navigation VLA models with two complementary analyses. 
First, attribution-based vision-action (VxA) analysis measures whether visual attribution localizes the named landmark and whether such localization is more frequent for acceptable local actions. 
\rev{Only two} of 60 VxA evaluations support a positive vision-action association.
Second, matched visual intervention compares the output change caused by editing the landmark region with that caused by applying the same edit to a comparable non-landmark region. To distinguish landmark-specific effects from general visual responsiveness, we also apply a full-frame edit that removes image structure from all observations in the evaluated temporal scope. 
\rev{All 23 configurations show} a positive mean response to the full-frame edit, but only six support a larger response to the landmark than to the matched control.
These results show that acceptable local actions and broad visual sensitivity are not, by themselves, reliable evidence that \rev{navigation VLA models rely} on the instructed landmarks.
\rev{Our project homepage is
\href{https://realjoshqsun.github.io/landmark-vln/}{here}.}
\end{abstract}

\section{Introduction}
\label{sec:intro}

\begin{figure*}[t!]
\centering
\begin{tikzpicture}[x=1cm,y=1cm]
  \path[use as bounding box] (0,1.10) rectangle (17.2,5.35);
  \node[anchor=north,align=center,text width=16.8cm,font=\scriptsize]
    at (8.60,5.25) {\textbf{Instruction:} Turn slight left and move towards
    the black chair.\\\textbf{Landmark:} black chair};

  \draw[black!14] (7.05,1.20)--(7.05,4.55);

  \node[anchor=north,align=center,font=\scriptsize\bfseries] at (1.95,4.45)
    {Correct action \cmark\\Correct focus \cmark};
  \node[anchor=north,inner sep=0pt] at (1.95,3.72)
    {\includegraphics[width=2.90cm]{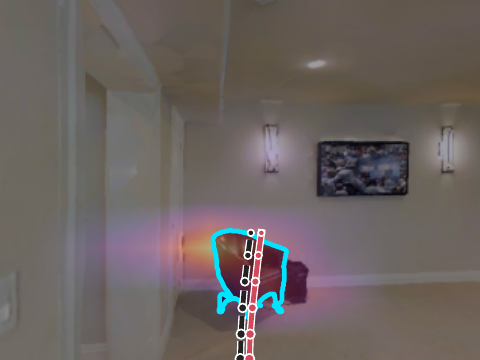}};

  \node[anchor=north,align=center,font=\scriptsize\bfseries] at (5.07,4.45)
    {Correct action \cmark\\Wrong focus \xmark};
  \node[anchor=north,inner sep=0pt] at (5.07,3.72)
    {\includegraphics[width=2.90cm]{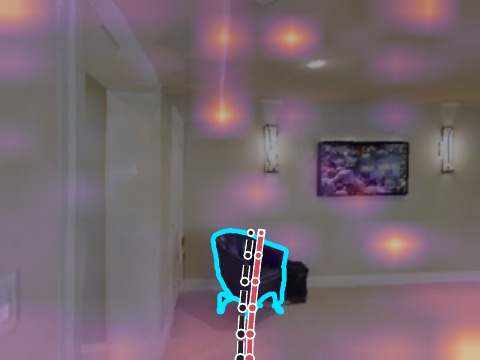}};

  \node[anchor=north,align=center,text width=9.6cm,font=\scriptsize\bfseries]
    at (12.13,4.45) {Which edit moves the prediction farther from the
    reference action?\\Landmark mask or same-shaped mask?};

  \node[anchor=north,font=\scriptsize\bfseries] at (9.68,3.65)
    {Landmark edit changes more \cmark};
  \node[anchor=north west,inner sep=0pt] at (7.45,3.20)
    {\includegraphics[width=2.20cm]{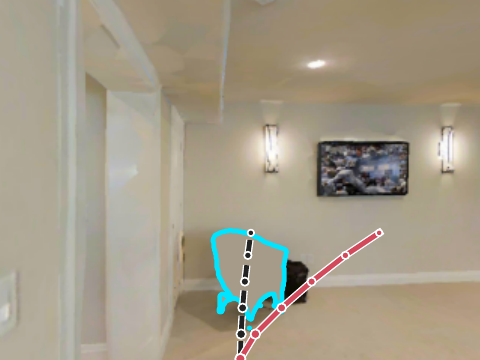}};
  \node[anchor=north west,inner sep=0pt] at (9.70,3.20)
    {\includegraphics[width=2.20cm]{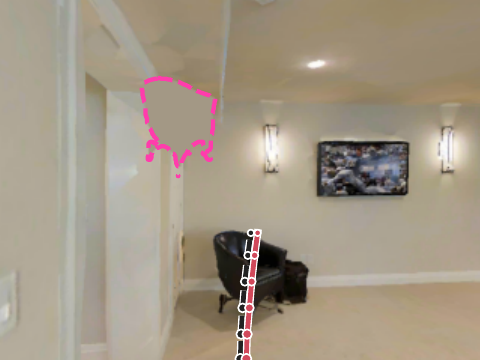}};

  \node[anchor=north,font=\scriptsize\bfseries] at (14.48,3.65)
    {Both edits change similarly ?};
  \node[anchor=north west,inner sep=0pt] at (12.25,3.20)
    {\includegraphics[width=2.20cm]{figures/teaser_findings_target_large.pdf}};
  \node[anchor=north west,inner sep=0pt] at (14.50,3.20)
    {\includegraphics[width=2.20cm]{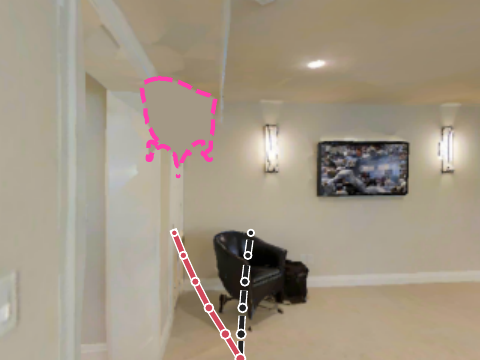}};
\end{tikzpicture}
\caption{\textbf{Left:} The observation is overlaid with the model's attention
heatmap, predicted action in red, and reference action in black. Both
predictions are correct, but the right heatmap fails to focus on the landmark.
\textbf{Right:} We apply per-channel mean edits separately to the landmark
mask in cyan and a same-shaped mask in magenta, then compare how far each
edited prediction moves from the reference action. A larger change under the
landmark edit indicates landmark-specific sensitivity. Similar changes give
an inconclusive comparison.}
\label{fig:teaser-motivation-findings-wide}
\end{figure*}

Vision-and-language navigation (VLN) asks an embodied agent to follow a
natural-language route instruction using a stream of egocentric
observations~\cite{anderson2018r2r,ku2020rxr,krantz2020vlnce}. Given an
instruction such as \emph{walk past the sofa, turn left at the piano, and stop
beside the round table}, the named landmarks connect successive language
segments to visual evidence along the route. Following the route requires the
agent to understand the instruction and its landmarks, match \rev{them}
to changing observations, and \rev{select actions}.

Recent navigation vision-language-action (VLA) models adapt pretrained
vision-language models to generate discrete or continuous actions~\cite{zhang2024navid,wei2025streamvln}. Standard metrics such as
success rate (SR) and nDTW \rev{measure navigation performance}
\cite{anderson2018r2r,ilharco2019ndtw}. \rev{Route-level success reflects many
decisions and other factors throughout an episode. We complement these metrics
with local action agreement at fixed, annotated navigation states to relate
visual landmark evidence directly to the current decision.} However, an acceptable local action can also
arise from recurring route patterns or language
regularities in the benchmark, without locating or interpreting the named
landmark. Such shortcut-based success does not ensure reliable action selection at decision points where the instructed landmark is needed to distinguish among otherwise plausible actions. \emph{Standard navigation metrics therefore do not characterize landmark-vision
alignment, and prior VLN works have not directly tested
instructed-landmark localization or landmark-specific output sensitivity.}

Evaluating this alignment requires a common unit that connects language,
visual evidence, and local behaviour. Navigation also introduces a temporal
requirement because a landmark may be visible now or may have appeared earlier
along the route. \rev{We construct \benchmark{} from R2R and RxR. Its 7,228 samples link} a navigation
instruction and a \rev{landmark phrase naming one physical object} with current and historical
observations, temporally aligned masks of the same physical landmark, and local
reference actions. 

\rev{We audit each model through attribution and intervention.
Attribution-based vision-action (VxA) analysis tests whether visual attribution
localizes the instructed landmark and whether localization is more frequent
for acceptable local actions, assessed against reference actions. Matched
visual intervention measures broad visual responsiveness through full-frame
edits in the selected temporal scope, then compares output changes from
editing the landmark and a same-shaped control region elsewhere to measure
landmark-specific output sensitivity. We also ask whether stronger landmark
attribution accompanies a larger landmark-specific intervention response.
These two views follow existing visual-explanation research
\cite{selvaraju2019hint,ying2022visfis,reich2023fpvg}.}

We apply the audit to eight released navigation VLA models:
OneVLA~\cite{zhang2026onevla},
InternVLA-N1~\cite{wei2025dualvln},
OmniVLA~\cite{hirose2025omnivla},
Uni-NaVid~\cite{zhang2024uninavid},
FantasyVLN~\cite{zuo2026fantasyvln},
StreamVLN~\cite{wei2025streamvln},
VAMOS~\cite{castro2025vamos}, and
TIC-VLA~\cite{huang2026ticvla}.
They span current-only and history-aware visual inputs as well as discrete
and continuous navigation outputs. The VxA analysis supports a positive
association between landmark localization and acceptable actions in two of 60
panels. Full-frame edits produce a positive mean response in all 23 configurations, whereas matched interventions support a larger response to the landmark than to its matched control in only six. 

Our contributions are:
\begin{itemize}
\item We introduce \benchmark{}, a 7,228-sample evaluation dataset that links
navigation instructions and singular landmark phrases to current and historical
observations, aligned same-instance masks, and local reference
actions.
\item \rev{Our framework reports} local action agreement,
landmark localization, and landmark-specific output sensitivity.
\item We evaluate eight released navigation VLA models on the same samples and
\textbf{show that binary localization rarely tracks acceptable actions, and 
landmark-specific sensitivity is supported in only a subset of configurations}.
\end{itemize}

\section{Related Work}
\label{sec:related}

\paragraph{VLN tasks and evaluation.}
R2R \cite{anderson2018r2r} introduced route following in Matterport environments, RxR \cite{ku2020rxr} expanded the
language and route diversity, and VLN-CE replaced graph teleportation with
continuous low-level control
\cite{krantz2020vlnce,hong2022bridging,hong2021vlnbert,krantz2021waypoint,11419772}. Standard metrics such as
success rate (SR) and nDTW measure navigation behaviour
\cite{anderson2018r2r,jain2019cls,ilharco2019ndtw}, but they do not reveal whether
an action uses the landmarks.

\paragraph{Navigation VLA models.}
Navigation VLAs adapt pretrained vision-language models through post-training
on navigation instruction-observation-action data. Recent systems report strong
SR across VLN benchmarks
\cite{zhang2024navid,zhang2024uninavid,wei2025streamvln,
hirose2025omnivla,castro2025vamos,cheng2025navila,yu2025correctnav}. However, none of them evaluate whether the model connects the observations to the
corresponding landmark.

\paragraph{Grounding and intervention tests in VQA.}
VQA routinely evaluates localization by comparing attention or gradient-based
importance with human-annotated visual evidence
\cite{das2016human,selvaraju2019hint,ying2022visfis,reich2023fpvg,xing2025large}.
Complementary intervention tests retain or mask selected visual regions and
measure whether the answer is preserved or changes
\cite{chen2020css,ying2022visfis,reich2023fpvg}. Localization and intervention
thus provide two established views of visual reliance in VQA. Our VxA and
matched intervention analyses adapt this template to sequential navigation actions,
temporal scopes, and landmark masks tracked across time.
\rev{NavBind examines instruction-action binding at navigation route
forks~\cite{sun2026navbind}.}

\section{Constructing \benchmark}
\label{sec:landmarkvln}

Both analyses require a common evaluation unit that connects one named
landmark to local behaviour and visual evidence. We therefore construct each
\benchmark{} sample by pairing a navigation instruction and singular landmark
phrase with a current RGB observation, its recent observation history,
temporally aligned masks of the same physical landmark, and a local
expert-action horizon.

\rev{We use} the continuous-environment releases of R2R and
RxR~\cite{anderson2018r2r,ku2020rxr,krantz2020vlnce}, restricting RxR to English.
FGR2R and Landmark-RxR provide \rev{fine-grained route supervision for
initial filtering and alignment}~\cite{hong2020fgr2r,he2021landmarkrxr}.
FGR2R aligns R2R sub-instructions with reference subpaths, while Landmark-RxR
provides aligned sub-instruction-subtrajectory units and route-critical points
and headings for RxR.

From this aligned base, we construct \benchmark{} in three stages.
\emph{Text only} selects phrases that can denote one physical
intervention target. \emph{Image-text, current} establishes that this target
is present, unique, and geometrically usable in the current observation.
\emph{Image-text, temporal} resolves landmark visibility in every history
frame, requires at least one historical localization, and verifies the
same-instance identity through time. It then yields samples for analysis.
Table~\ref{tab:construction-scale}
summarizes the scale at each stage.

\begin{table}[t]
\centering\small
\caption{Construction scale by stage. The text-only stage selects singular physical
intervention targets. The current-view stage verifies visible, unique, and
geometrically usable grounding. The temporal stage resolves visibility across
history and verifies the same-instance identity through time. The table reports each stage in its
appropriate unit. Pairing the 118,706 text-selected identities with a few candidate frames each produces the 350,660 candidate-frame pairs used
by the current-view stage.}

\label{tab:construction-scale}
\begin{tabularx}{\linewidth}{lXX}
\toprule
Stage & Input & Output \\
\midrule
Text only & 39,356 English instructions\rev{,} 374,270 route-local mentions &
118,706 singular candidate identities \\
\begin{tabular}[t]{@{}l@{}}
Image-text,\\
current
\end{tabular}
& 350,660 candidate-frame pairs &
19,742 current-grounded identities \\
\begin{tabular}[t]{@{}l@{}}
Image-text,\\
temporal
\end{tabular}
& 19,742 current-grounded identities &
\textbf{7,228} intervention-ready samples \\
\bottomrule
\end{tabularx}
\end{table}

\subsection{Text-only selection}
\label{sec:text-filter}

The text-only stage screens landmark mentions in the navigation instructions to
retain phrases that can denote one physical object instance and thus serve as
candidate intervention targets. It relies exclusively on the instruction text,
without consulting RGB observations. The rules below \rev{exclude phrases that
do not identify one physical object}.

\paragraph{Object instance rather than region or surface.}
\emph{The round table} is retained, whereas \emph{the kitchen},
\emph{the left side of the room}, and \emph{the floor} are rejected. These
phrases describe regions or surfaces rather than localized objects.

\paragraph{One resolvable object instance.}
A candidate phrase must resolve from the instruction text to exactly one
physical object instance, and it can fail this requirement in three ways.
First, an instruction such as \emph{Walk past the chair or the table and enter
the hallway} leaves two alternative referents, i.e., \emph{the chair or the table}. Second, \emph{the chair} is rejected when multiple chair mentions in the
instruction remain possible referents and no linguistic cue selects one. Third,
explicit counts and plurals, such as \emph{two lamps} and \emph{the chairs},
denote multiple instances and are rejected.

\paragraph{Represented versus representing object.}
For example, in \emph{Walk past the picture of a boat and turn left}, the
physical scene object is the picture, not the depicted boat. The referent is
therefore assigned to \emph{the picture}, since the depicted boat has no
corresponding 3-D scene instance to mask.

\subsection{Current-View Landmark Grounding}
\label{sec:visual-filter}

The text stage yields a route-local sub-instruction and a singular landmark
phrase, but not its visual evidence. The goal here is to determine whether the
landmark is visible and uniquely identifiable in the corresponding current RGB
observation and, if so, to obtain its landmark mask.

\rev{We construct masks through} candidate generation, semantic verification, and geometric
verification.

\paragraph{\rev{Candidate generation.}}
SAM3~\cite{carion2026sam3} is a text-prompted
segmenter that produces open-vocabulary masks conditioned on the landmark
phrase. In parallel, YOLOE~\cite{wang2025yoloe}
is an open-vocabulary detector and segmenter that we run with a fixed,
prompt-free vocabulary, so that its object inventory is independent of the
target phrase. For \emph{the black chair}, SAM3 may propose a chair-shaped
mask while YOLOE independently lists the visible objects. Their outputs
form the candidate set passed to semantic verification.

\paragraph{\rev{Semantic verification.}}
We then combine the YOLOE inventory with
Florence-2~\cite{xiao2024florence2}, a vision-language model used for
prompt-free object detection and region description, and
Qwen2.5-VL~\cite{bai2025qwen25vl}, a multimodal language model used to assess
target presence and enumerate visible objects. For the same black-chair
example, Florence-2 describes the proposed region as
a chair and Qwen2.5-VL reports one visible black chair. This allows the same
SAM3 candidate to proceed. A candidate is rejected when the target is
unsupported or its correspondence is ambiguous.

\paragraph{\rev{Geometric verification.}}
Semantic survivors are finally checked using
GroundingDINO~\cite{liu2024groundingdino} and
OWLv2~\cite{minderer2023scaling}, two text-conditioned open-vocabulary
detectors that independently localize the landmark with bounding boxes.
SAM2.1~\cite{ravi2025sam2}, a
class-agnostic promptable segmenter, converts these boxes into masks. The
resulting masks must identify the same object as the SAM3 proposal. Otherwise,
the candidate is rejected. The corroborated SAM3 proposal becomes the final
landmark mask.

\paragraph{Visual uniqueness.}
If a frame contains several chairs and the phrase does not determine which
one is route-relevant, the pair is rejected. A chair is retained only when one
individual can be isolated in the current frame.

\subsection{Temporal Landmark Verification}
\label{sec:temporal-filter}

Current-view grounding establishes that a landmark is visible at the decision
frame, but some history-based models also require reliable evidence about
its earlier visibility and identity. These baselines use four
history observations in addition to the current observation, so each
candidate must provide a complete five-observation sequence. Of the 19,742
current-grounded identities, 17,813 satisfy this requirement.

We reuse the current-view grounding models to assess each history frame. A
frame is \texttt{resolved} when their evidence either yields a reliable landmark
localization or consistently indicates that the landmark is not visible.
Otherwise, it is \texttt{unresolved}. We retain only sequences in which all
four history frames are \texttt{resolved} and the landmark is localized in at
least one of them. For localized frames, GroundingDINO and OWLv2 independently produce
phrase-conditioned boxes, which SAM2.1 converts into masks\rev{. Inconsistent}
localizations are rejected. SAM2.1 is then initialized with the verified
current-frame mask and run backward through the history to reject identity
switches between similar objects. We conducted a human audit of 100 samples.

\begin{table}[t]
\centering\small
\caption{\rev{Native action spaces of the evaluated models.}
Discrete outputs are \texttt{FORWARD}, \rev{\texttt{LEFT}, \texttt{RIGHT}},
or \texttt{STOP}. \rev{Turns use $30^\circ$ for Uni-NaVid and FantasyVLN and
$15^\circ$ for the other discrete models.}
For \emph{continuous} outputs, $N\!\times\!D$ denotes
\rev{$N$ waypoints of dimension $D$.} }
\label{tab:models}
\begin{tabularx}{\linewidth}{@{}llX@{}}
\toprule
Model & Venue / year & Action type and horizon \\
\midrule
Uni-NaVid~\cite{zhang2024uninavid} & RSS 2025 & Discrete, $\leq4$ steps \\
VAMOS~\cite{castro2025vamos} & arXiv 2025 & Continuous, $5\!\times\!2$ waypoints \\
OmniVLA~\cite{hirose2025omnivla} & ICRA 2026 & Continuous, $8\!\times\!4$ waypoints \\
StreamVLN~\cite{wei2025streamvln} & ICRA 2026 & Discrete, $4$ steps \\
InternVLA-N1~\cite{wei2025dualvln} & ICLR 2026 & Discrete, $\leq4$ steps\rev{.} Continuous, $33\!\times\!2$ waypoints \\
FantasyVLN~\cite{zuo2026fantasyvln} & CVPR 2026 & Discrete, $\leq5$ steps \\
TIC-VLA~\cite{huang2026ticvla} & ICML 2026 & Continuous, $30\!\times\!2$ waypoints \\
OneVLA~\cite{zhang2026onevla} & arXiv 2026 & Discrete, $5$ steps \\
\bottomrule
\end{tabularx}
\end{table}

\begin{figure}[t!]
\centering
\includegraphics[width=\linewidth]{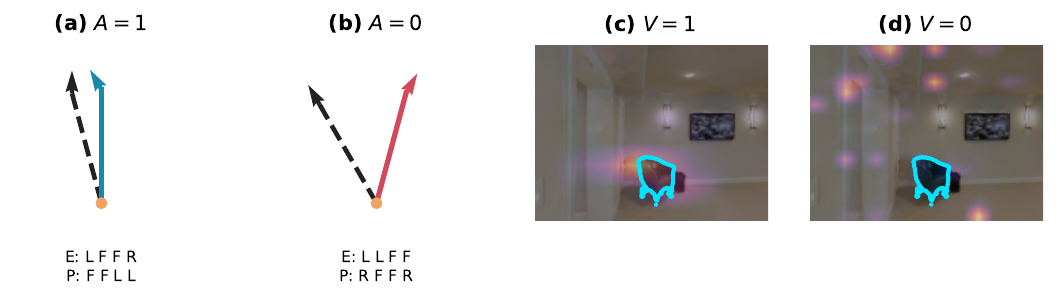}
\caption{Operational examples of $A$ and $V$. Panels (a,b) show two discrete
action-diagnostic outcomes, and panels (c,d) show the two visual-localization
outcomes. Dashed black trajectories are expert references, colored trajectories
are model predictions, and E/P list the corresponding actions. Cyan outlines
mark target landmarks. Heatmap intensity is normalized within each panel.}
\label{fig:av-labels}
\end{figure}

\section{Alignment Analysis}
\label{sec:protocol}

\rev{Across 7,228 samples, we audit local action acceptability $A$, landmark
attribution localization $V$, and matched intervention response $E$.
VxA relates $V$ to $A$ (Sec.~\ref{sec:vxa}). Matched intervention compares target
and control edits (Sec.~\ref{sec:causal}). We compare continuous attribution
with $E$.}

\subsection{Temporal scopes}
\label{sec:inputs}

OneVLA, InternVLA-N1, Uni-NaVid, FantasyVLN, StreamVLN, and TIC-VLA consume
both current and historical visual observations, so we analyze their current
(\C{}), memory (\M{}), and joint current-memory (\CM{}) scopes. OmniVLA and
VAMOS depend only on the current frame, so only \C{} is analyzed. Table~\ref{tab:models} describes the
action spaces of the models.
\rev{For attribution, the scopes select tokens to analyze while retaining each
model's normal visual input. For intervention, they select observations to edit.}

\subsection{Attribution-based vision-action analysis}
\label{sec:vxa}

\paragraph{Action variable.}
\rev{To define $A$, we convert model predictions to discrete navigation actions
and compare their resulting position, heading, action order, and collisions
with the local reference. Predicted and reference \texttt{STOP} status must
agree. For VAMOS and TIC-VLA, we quantize the first predicted waypoint and
compare it with the first reference action. We set $A=1$ when the acceptance
criterion is met and $A=0$ otherwise.}
Figure~\ref{fig:av-labels}(a,b) shows two examples\rev{.}

\paragraph{Visual variable.}
The binary visual diagnostic $V$ evaluates whether the model's
visual attribution is localized on the target landmark. For the
current scope, following region-mass localization metrics
\cite{arras2022clevrxai}, we sum attribution mass inside the current target mask. For the
memory and joint scopes, we sum attribution mass inside the corresponding
target masks across the active frames. \rev{We compare this mass with masses
under valid translations of the same mask and set $V=1$ when its empirical
upper-tail spatial rank is at most .05 and $V=0$ otherwise.
Localization uses multiple translated masks, whereas intervention uses one
deterministic matched control.}
Figure~\ref{fig:av-labels}(c,d) shows the two visual outcomes.

\paragraph{VxA outcome.}
Together, $V$ and $A$ define four outcomes. $(V,A)=(1,1)$ combines an
acceptable action with localized landmark attribution. $(0,1)$ is an
acceptable action without detected landmark localization. $(1,0)$ combines
an action-diagnostic failure with localized landmark attribution. $(0,0)$
means that both diagnostics are negative.

\paragraph{Attribution estimators.}
We evaluate attention, gradient, and attention$\times$gradient for visual attribution.
\rev{Depending on the model, the differentiable action score $s$ uses clean-output
token log-probabilities, selected navigation-output components, or inner products
of waypoint outputs with a detached clean-output reference.}
Let $a_{\ell,h,q,v}$ denote the attention from action-related query $q$ to
visual token $v$ at layer $\ell$ and head $h$.
\emph{Attention} aggregates how strongly the action-related queries attend to
that token \cite{jain2019attention,wiegreffe2019attention}.
\rev{For all models except TIC-VLA, \emph{gradient} aggregates the absolute
sensitivity of $s$ to post-softmax attention probabilities
\cite{selvaraju2017gradcam}. These attention-probability-based constructions use}
\begin{equation}
\begin{aligned}
H_{\mathrm{att}}(v)
&\propto \sum_{\ell}\operatorname{mean}_{h,q}a_{\ell,h,q,v}, \\
H_{\mathrm{grad}}(v)
&\propto \sum_{\ell}\operatorname{mean}_{h,q}
\left\lvert\frac{\partial s}{\partial a_{\ell,h,q,v}}\right\rvert.
\end{aligned}
\end{equation}
\rev{For these seven models, \emph{attention$\times$gradient} multiplies each
attention probability by its corresponding absolute gradient before averaging
over heads and queries and aggregating layers \cite{chefer2021transformer}.
TIC-VLA instead differentiates $s$ with respect to projected source-token
features. It averages absolute feature gradients over channels, then multiplies
this token-level score by aggregated attention.}

\paragraph{Temporal operations.}
\rev{For \C{}, attribution is restricted to tokens from the current observation
and compared with its landmark mask. For \M{}, it is restricted to tokens from
the four history observations and compared with their masks. For \CM{}, both
current and history observations contribute. Attribution is normalized over the
active visual scope.}

\paragraph{Reported measures.}
\rev{We report localization rate $\pi_V=\Pr(V=1)$ and vision-action risk
difference $\mathrm{RD}=\Pr(V=1\mid A=1)-\Pr(V=1\mid A=0)$.
Table~\ref{tab:vxa} pairs $\widehat{\pi}_V\,(\%)\,|\,\widehat{\mathrm{RD}}$.
Thus, 32.4\,$|$\,-.023 indicates 32.4\% localization, 2.3 percentage points less
frequent for $A=1$ than $A=0$.}

\subsection{Matched visual intervention analysis}
\label{sec:causal}

\begin{figure}[t]
\centering
\includegraphics[width=\linewidth]{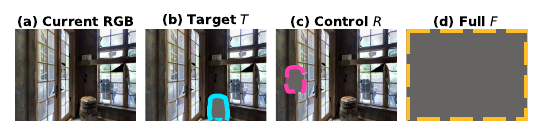}
\caption{Intervention regions in the current observation.
(a) Current RGB, (b) cyan target mask $T$ on the named landmark,
(c) dashed magenta same-shaped, non-overlapping control $R$, and (d)
full-frame support $F$. In (b-d), pixels inside the indicated support
are replaced by the per-channel mean. The target and control define the
matched effect $E=D^T-D^R$, while the full-frame condition measures broad
visual sensitivity.}
\label{fig:causal-masks}
\end{figure}

Matched visual intervention measures landmark-specific output sensitivity. It
tests whether the model's native navigation output responds more to editing
the instructed landmark than to applying the same edit at another image
location.

\paragraph{Intervention regions.}
The analysis uses three visual supports (Figure~\ref{fig:causal-masks}). The
\emph{target mask} $T$ covers the instruction-named landmark. The
\emph{matched-control mask} $R$ is a deterministic translation of the same
mask to a location whose bounding box does not overlap $T$. Target and control
therefore contain the same number and arrangement of pixels at different
locations. \rev{Mean-color replacement is an established image-perturbation
operation \cite{fong2017meaningful,hooker2019benchmark,maini2024tmars}.
We replace pixels inside each selected mask with the rounded per-channel mean
of the pixels outside that mask in the same frame. Applied to $T$, this replaces
the landmark's interior pixels while potentially retaining its outline and
surrounding context. For $R$, it edits an equally shaped region elsewhere.
Target and control use the same edit rule, with the clean input as their shared
baseline.} The
\emph{full-frame support} $F$ covers every pixel in the active observation or
observations. \rev{Its edit replaces each active frame with its rounded per-channel mean
to measure broad visual sensitivity within the selected temporal scope.
The matched effect compares $T$ and $R$.}

\paragraph{Temporal intervention scopes.}
A \C{} intervention edits the selected support in the current observation and
keeps the history unaltered. An \M{} intervention edits the selected
support in all four history observations and keeps the current RGB clean. A
\rev{\CM{} intervention edits current and history.}

\paragraph{Matched landmark-specific effect.}
For each intervention, we run the model on the edited input and compare its
output with the clean output using a model-specific response measure $d_p$.
Depending on the model's output format, $d_p$ measures a waypoint or
action-vector distance, an action-sequence distance, or the decrease in log
probability assigned to the clean prediction.

Let $f_p(X_i)$ denote model $p$'s clean navigation output for sample $i$.
For temporal scope $s\in\{\C{},\M{},\CM{}\}$ and support
$Q\in\{T,R\}$, define
\begin{align}
 D_{i,s}^{Q,(p)}
 &=d_p\!\left(f_p(X_i),f_p(\mathcal I(X_i,Q_{i,s}))\right),\\
 E_{i,s}^{(p)}
 &=D_{i,s}^{T,(p)}-D_{i,s}^{R,(p)}.
 \label{eq:causal-effect}
\end{align}
The effect $E_{i,s}^{(p)}$ compares the response to editing the
landmark with the response to editing its matched control. We report the mean
$\bar E_s^{(p)}$ over eligible samples. A positive mean indicates greater
sensitivity to the landmark than to the control. Table~\ref{tab:causal}
reports these means.

\paragraph{Attribution-intervention agreement.}
For this comparison, we use attention$\times$gradient and define the
continuous attribution score $\Delta m$ as the target
attribution mass minus the mean mass over the translated spatial
controls used by $V$. Within each model, we compute the Spearman correlation between $\Delta m$ and the
matched target-control effect $E$ over jointly valid samples. This asks
whether samples with a stronger attribution preference for the
landmark also show a larger landmark-specific output response.

\section{Evaluation Design}
\label{sec:experiments}

\paragraph{Analysis grid.}
We evaluate all eight models on the same 7,228 samples.
\rev{Both analyses use the full navigation instruction and the same
observations, formatted through each model's released input interface.}
Their available visual
inputs define 20 model-scope combinations. The six temporal models contribute
\C{}, \M{}, and \CM{}, while the two current-only models contribute \C{}.
\rev{For VxA, every combination uses attention, gradient, and
attention$\times$gradient, giving 60 panels. We report each panel's localization
rate and risk difference.}

The matched intervention analysis keeps the full instruction fixed and uses
the same 20 model-scope combinations. InternVLA-N1 exposes continuous and
discrete response branches, which are evaluated separately. This produces 23
model-branch-scope estimates.

\FloatBarrier
\section{Results}
\label{sec:results}

The results answer three linked questions. \textbf{Q1}: Does landmark localization
occur more often when a model produces an acceptable action? \textbf{Q2}: Does a
model respond more to editing the instructed landmark than to editing a
matched control? \textbf{Q3}: Do continuous attribution and intervention response
vary together across samples? 

\begin{figure*}[t]
\centering
\includegraphics[width=.94\textwidth]{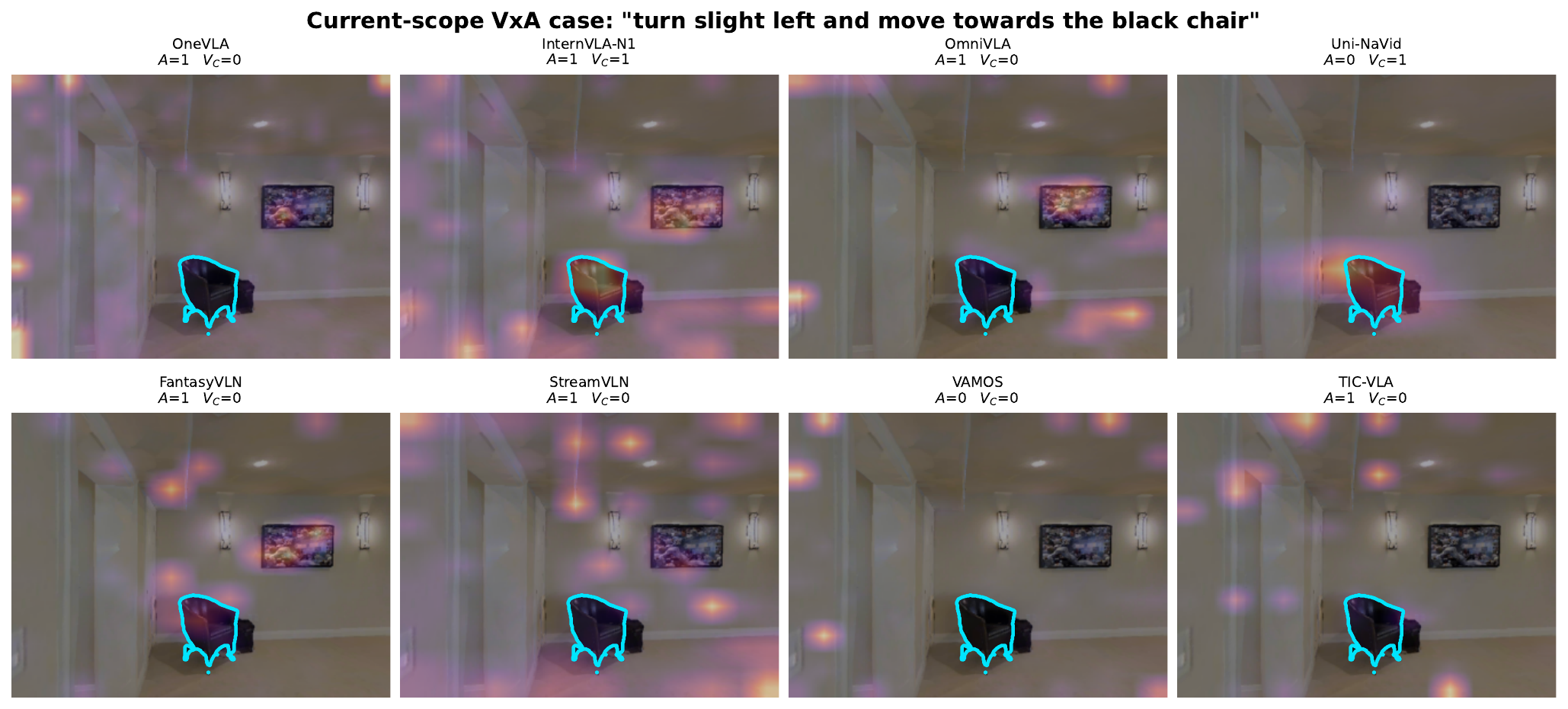}
\caption{VxA case study for \emph{turn slight left and move towards the
black chair}. Panels overlay current-scope attention$\times$gradient
attribution. Cyan outlines the target. Labels give the action diagnostic $A$
and localization decision $V_{\C{}}$. Together, the panels show all four
combinations of the two diagnostics. Heatmaps are normalized independently
within each model.}
\label{fig:vxa-case}
\end{figure*}

\begin{figure*}[t]
\centering
\includegraphics[width=.9\textwidth]{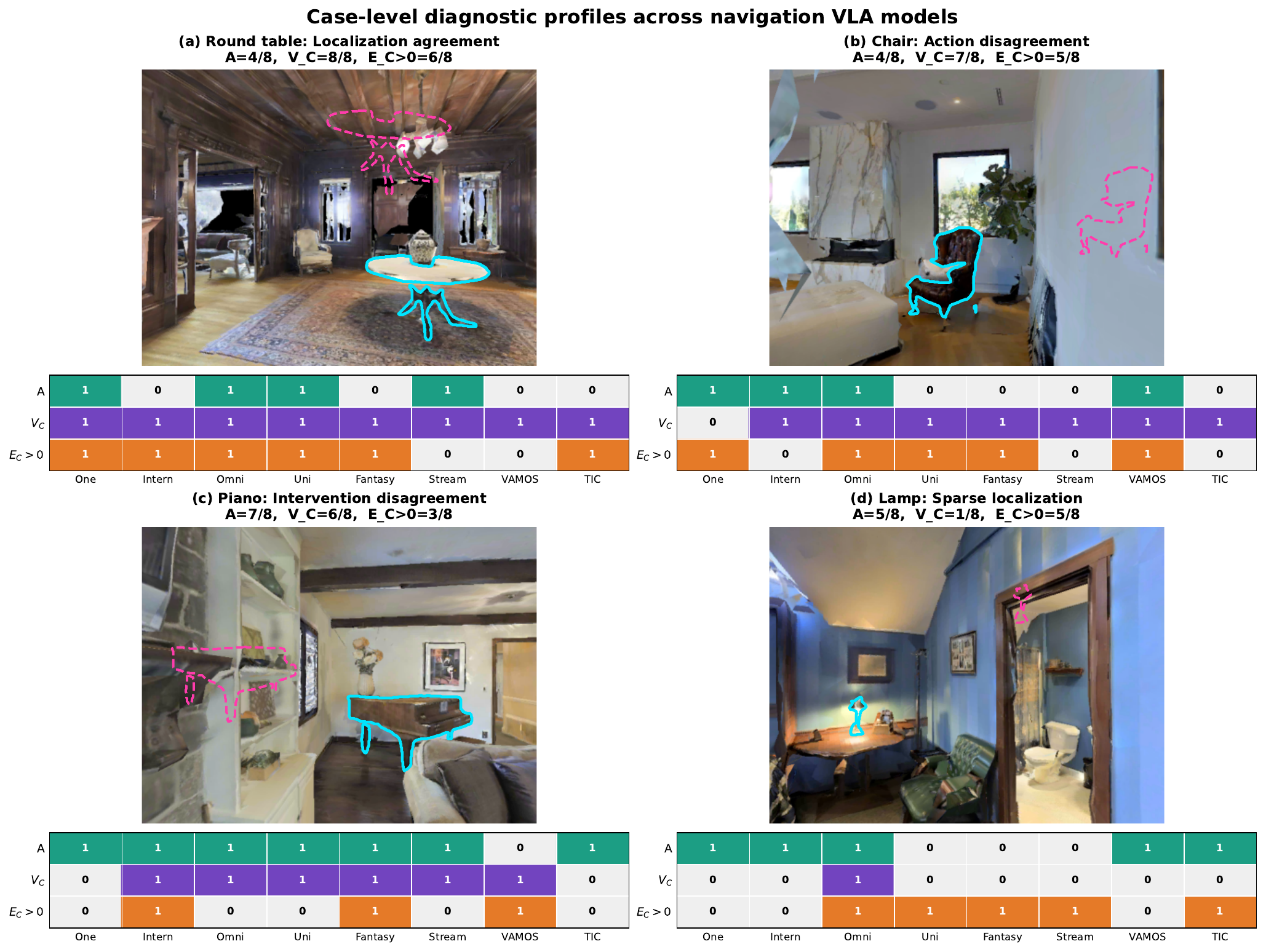}
\caption{Case-level diagnostic profiles for four selected \benchmark{}
samples: (a) round table, (b) chair, (c) piano, and (d) lamp. Cyan is the
instruction target and dashed magenta is the same-shaped control. Each panel
reports the local action diagnostic $A$, the binary current-scope
attention$\times$gradient result $V_{\C{}}$, and whether $E_{\C{}}>0$ for all eight
models. Here $E_{\C{}}>0$ means that the target edit produces a larger native
response than the matched control for that sample. The InternVLA-N1 column
uses its discrete response branch for $E_{\C{}}$.}
\label{fig:selected-profiles}
\end{figure*}

\subsection{Attribution-Based Vision-Action Analysis}
\label{sec:results-vxa}

\begin{table}[!t]
\centering
\scriptsize
\setlength{\tabcolsep}{1.0pt}
\renewcommand{\arraystretch}{.82}
\caption{\rev{Landmark-localization rate (\%) $|$ risk difference for
full instructions. $\dagger$ marks supported negative associations
under within-model Holm-adjusted two-sided
Fisher tests ($p_{\mathrm H}<.05$). Dashes indicate unavailable scopes.}}
\label{tab:vxa}
\resizebox{\columnwidth}{!}{%
\begin{tabular}{@{}ll
  r@{\,$|$\,}l
  r@{\,$|$\,}l
  r@{\,$|$\,}l@{}}
\toprule
Model & Estimator
& \multicolumn{2}{c}{\C{}}
& \multicolumn{2}{c}{\M{}}
& \multicolumn{2}{c}{\CM{}} \\
\midrule
\multirow{3}{*}{OneVLA}
 & Attention & 42.5 & -.008 & 45.4 & +.024 & 47.6 & +.009 \\
 & Gradient & 28.8 & \rev{-.038\textsuperscript{\(\dagger\)}} & 21.5 & +.008 & 25.3 & -.012 \\
 & Attention$\times$gradient & 32.4 & -.023 & 25.3 & -.002 & 29.1 & -.023 \\
\midrule
\multirow{3}{*}{InternVLA-N1}
 & Attention & 56.0 & +.004 & 52.3 & -.012 & 57.3 & -.006 \\
 & Gradient & 49.9 & -.018 & 64.2 & -.008 & 65.4 & -.018 \\
 & Attention$\times$gradient & 56.2 & +.004 & 52.2 & -.011 & 57.5 & -.008 \\
\midrule
\multirow{3}{*}{OmniVLA}
 & Attention & 51.7 & +.026 & \multicolumn{2}{c}{--} & \multicolumn{2}{c}{--} \\
 & Gradient & 13.8 & -.020 & \multicolumn{2}{c}{--} & \multicolumn{2}{c}{--} \\
 & Attention$\times$gradient & 51.4 & +.021 & \multicolumn{2}{c}{--} & \multicolumn{2}{c}{--} \\
\midrule
\multirow{3}{*}{Uni-NaVid}
 & Attention & 54.8 & \rev{-.034\textsuperscript{\(\dagger\)}} & 47.1 & \rev{-.039\textsuperscript{\(\dagger\)}} & 54.7 & \rev{-.035\textsuperscript{\(\dagger\)}} \\
 & Gradient & 39.6 & +.016 & 34.4 & +.014 & 40.3 & +.009 \\
 & Attention$\times$gradient & 55.1 & -.029 & 47.7 & -.030 & 54.8 & \rev{-.033\textsuperscript{\(\dagger\)}} \\
\midrule
\multirow{3}{*}{FantasyVLN}
 & Attention & 56.5 & -.005 & 54.6 & +.003 & 58.8 & -.0002 \\
 & Gradient & 35.8 & +.024 & 49.2 & +.009 & 50.6 & +.017 \\
 & Attention$\times$gradient & 53.1 & -.023 & 50.9 & -.032 & 55.3 & -.019 \\
\midrule
\multirow{3}{*}{StreamVLN}
 & Attention & 40.2 & +.015 & 45.5 & +.010 & 45.4 & +.006 \\
 & Gradient & 38.9 & \rev{+.046} & 55.4 & +.027 & 56.0 & \rev{+.041} \\
 & Attention$\times$gradient & 40.0 & +.017 & 44.3 & +.007 & 44.7 & +.001 \\
\midrule
\multirow{3}{*}{VAMOS}
 & Attention & 28.6 & +.006 & \multicolumn{2}{c}{--} & \multicolumn{2}{c}{--} \\
 & Gradient & 21.1 & +.006 & \multicolumn{2}{c}{--} & \multicolumn{2}{c}{--} \\
 & Attention$\times$gradient & 30.8 & -.006 & \multicolumn{2}{c}{--} & \multicolumn{2}{c}{--} \\
\midrule
\multirow{3}{*}{TIC-VLA}
 & Attention & 14.7 & +.004 & 52.4 & \rev{-.049\textsuperscript{\(\dagger\)}} & 19.7 & -.009 \\
 & Gradient & 12.4 & +.003 & 41.9 & -.028 & 18.5 & -.010 \\
 & Attention$\times$gradient & 11.4 & +.008 & 38.2 & -.022 & 13.4 & -.002 \\
\bottomrule
\end{tabular}%
}
\end{table}

\paragraph{Landmark localization rarely identifies acceptable actions.}
\rev{Across 60 panels, localization rates range from 11.4\% to 65.4\%, while
RD ranges from -.049 to +.046, with 28 positive and 32 negative estimates.
Only StreamVLN gradient in \C{} and \CM{} supports a positive association
under the model-wise Holm-adjusted Fisher tests.} \textbf{Frequent landmark
localization therefore rarely accompanies a supported relationship with action
acceptability.}

\paragraph{Different attribution methods give different VxA patterns.}
The three ways of computing attribution do not always give the same
relationship between landmark localization and action acceptability for a
fixed model and visual scope. For Uni-NaVid, gradient gives positive values in
\C{}, \M{}, and \CM{}, while attention and attention$\times$gradient give
negative values. StreamVLN gives positive values under all three methods and
in all three scopes, with the largest values under gradient: +.046 in \C{},
+.027 in \M{}, and +.041 in \CM{}. The attribution calculation therefore
changes both the direction and size of the measured relationship.

\subsection{Matched Visual Intervention Analysis}
\label{sec:results-intervention}

The matched intervention analysis asks whether a model responds more to
editing the instruction-named landmark than to editing its matched control.

\begin{table}[!t]
\centering\footnotesize
\setlength{\tabcolsep}{3pt}
\caption{Mean matched target-control response $\bar E_s$ \rev{over eligible
samples. Bold positive entries pass within-model/branch Holm-adjusted
one-sided trajectory sign-randomization tests ($p_{\mathrm H}<.05$).}}
\label{tab:causal}
\begin{tabular*}{\columnwidth}{@{\extracolsep{\fill}}lrrr@{}}
\toprule
Model & \C{} & \M{} & \CM{} \\
\midrule
OneVLA & -.0197 & \rev{\textbf{+.0176}} & -.0946 \\
InternVLA-N1 (cont.) & +.00181 & +.00595 & \rev{\textbf{+.00938}} \\
InternVLA-N1 (disc.) & -.0131 & -.0234 & -.0369 \\
OmniVLA & \rev{\textbf{+.00191}} & -- & -- \\
Uni-NaVid & +.0100 & -.0164 & -.0050 \\
FantasyVLN & -.0373 & +.1068 & -.5450 \\
StreamVLN & -.0654 & -.0041 & -.0784 \\
VAMOS & \rev{\textbf{+.00279}} & -- & -- \\
TIC-VLA & -.00454 & \rev{\textbf{+.01075}} & \rev{\textbf{+.00971}} \\
\bottomrule
\end{tabular*}
\end{table}

\paragraph{\rev{Broad visual response.}}
\label{par:results-broad-response}
\rev{Full-frame edits yield positive observed native-scale means in all 23
configurations. Estimates use native-comparable pairs, with only 31
for InternVLA-N1 continuous \CM{}.}

\paragraph{Matched edits isolate landmark-specific response.}
\rev{In Table~\ref{tab:causal}, positive means favor the target and negative
means favor the control. A nonpositive effect does not establish landmark
non-use: both regions can influence the output, and usable landmark cues may
remain.} \textbf{\rev{Under Holm correction across scopes within each model and
native output branch, 6 of 23 configurations support a larger mean response to
the landmark than to its matched control.}}

\subsection{Attribution-Intervention Association}
\label{sec:results-association}

\begin{table}[t]
\centering
\footnotesize
\setlength{\tabcolsep}{5pt}
\caption{\rev{Continuous landmark attribution and matched response. Each entry is Spearman $\rho$ between the attention$\times$gradient
score $\Delta m$ and $E$ over jointly valid samples per configuration.}}
\label{tab:association-summary}
\begin{tabular*}{\columnwidth}{@{\extracolsep{\fill}}lccc@{}}
\toprule
Model/branch & \C{} $\rho$ & \M{} $\rho$ & \CM{} $\rho$ \\
\midrule
OneVLA & .100 & .105 & .106 \\
InternVLA-N1 (cont.) & .087 & .119 & .144 \\
InternVLA-N1 (disc.) & .031 & .047 & .073 \\
OmniVLA & .182 & -- & -- \\
Uni-NaVid & .044 & .016 & .057 \\
\rev{FantasyVLN} & \rev{.035} & \rev{.034} & \rev{.046} \\
\rev{StreamVLN} & \rev{.045} & \rev{.065} & \rev{.033} \\
\rev{VAMOS} & \rev{.030} & \rev{--} & \rev{--} \\
\rev{TIC-VLA} & \rev{.118} & \rev{.020} & \rev{.024} \\
\bottomrule
\end{tabular*}
\end{table}

\paragraph{The two analyses have a weak positive association.}
Across the 23 configurations, Spearman correlations between \rev{the continuous
attention$\times$gradient score $\Delta m$} and the matched target-control response are all positive
but small, ranging from .016 to .182, with a median of .047
(Table~\ref{tab:association-summary}). OmniVLA \C{} has the largest
correlation at $\rho=.182$.

\paragraph{History does not consistently strengthen the association.}
The correlation increases from \C{} to \CM{} for InternVLA-N1 and Uni-NaVid,
remains nearly constant for OneVLA and FantasyVLN, and decreases for StreamVLN
and TIC-VLA. The weak positive association therefore appears across models,
while its variation across visual scopes is model-dependent.

\subsection{Case Study}
\label{sec:results-cases}

\rev{The cases illustrate how the diagnostics separate within individual
instructions and observations.}

\paragraph{All four VxA outcomes in one observation.}
\rev{Figure~\ref{fig:vxa-case} compares eight models on the same black-chair
observation. InternVLA-N1 passes both diagnostics. Uni-NaVid localizes the chair
but fails $A$. VAMOS fails both. The other five models pass $A$ without detected
localization. The observation therefore exhibits all four VxA outcomes.}

\paragraph{Joint diagnostic profiles.}
\rev{Figure~\ref{fig:selected-profiles} adds current-scope $E>0$ to $A$ and $V$
for four cases. The round table and chair show high localization but split
action acceptability. The piano combines frequent $A/V$ positives with fewer
positive matched effects. The lamp shows the opposite pattern. These selected
illustrations complement the aggregate results in Tables~\ref{tab:vxa}
and~\ref{tab:causal} and the continuous comparison in
Sec.~\ref{sec:results-association}.}

\FloatBarrier
\section{Conclusion}

\rev{We introduced \benchmark{} to audit navigation VLA models' reliance on
instructed landmarks.} Across eight models, acceptable local actions and broad
visual responses were not, by themselves, reliable evidence of
\rev{such reliance}.

\FloatBarrier
{
  \small
  \bibliographystyle{ieeenat_fullname}
  \bibliography{main}
}

\section*{Acknowledgments}

\rev{Resources used in preparing this research at the University of Waterloo
were provided by the province of Ontario and the government of Canada through
a Discovery Grant from the Natural Sciences and Engineering Research Council
of Canada (NSERC: \mbox{RGPIN-2025-07221}) and an AI4Design grant from the
National Research Council (NRC: \mbox{AI4D-121-1}).
This research was enabled in part by support provided by Compute Ontario and
the Digital Research Alliance of Canada.
This work also received partial support from the Zhejiang Provincial Natural
Science Foundation of China under Grant No.\ \mbox{LQN26F030037}.
The authors additionally thank the High Performance Computing Center at
Eastern Institute of Technology, Ningbo, and the Ningbo Institute of Digital
Twin for computational resources.}

\end{document}